\documentclass{bmvc2k}

\hypersetup{
  pdftitle={Re-calibrated Contrastive Loss for Transformation-Aware Prompt Conditioning in Vision-Language Models},
  pdfauthor={Seungmin Oh, Seunghun Kang, Jongbin Ryu},
  pdfsubject={Contrastive learning objective function for efficient transfer learning},
  pdfkeywords={vision-language models, transfer learning, contrastive loss, prompt conditioning},
}

\usepackage[utf8]{inputenc}
\usepackage[T1]{fontenc}
\usepackage{url, hyperref}
\usepackage{booktabs, multirow, colortbl}
\usepackage{microtype}
\usepackage{xcolor}
\usepackage{amsmath, amsfonts, amssymb, nicefrac, bm}
\usepackage[capitalize,noabbrev]{cleveref}

\usepackage{listings}
\usepackage{setspace}
\usepackage{inconsolata}
\definecolor{pygKw}{rgb}{0.00,0.50,0.00}
\definecolor{pygCom}{rgb}{0.24,0.48,0.48}
\definecolor{pygStr}{rgb}{0.73,0.13,0.13}
\definecolor{pygFun}{rgb}{0.00,0.00,1.00}
\lstdefinestyle{alg}{
  language=Python,
  deletekeywords={[2]sum},
  basicstyle=\ttfamily\footnotesize\setstretch{1.2},
  keywordstyle=\bfseries\color{pygKw},
  commentstyle=\color{pygCom},
  stringstyle=\color{pygStr},
  moredelim=[s][\color{pygStr}]{"""}{"""},
  emph={recal,loss,loss_t},emphstyle=\color{pygFun},
  showstringspaces=false,
  columns=fullflexible, keepspaces=false,
  framesep=2mm, xleftmargin=0pt, aboveskip=0pt, belowskip=0pt,
}

\newcommand{\hquad}{\hspace{0.5em}}
\newcommand{\qqquad}{\hspace{0.25em}}
\newcommand{\Ip}{I$'$}
\newcommand{\Tp}{T$'$}
\newcommand\itmark{\raisebox{-0.2em}{\includegraphics[width=0.95em]{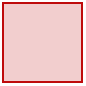}}}
\newcommand\iptpmark{\raisebox{-0.2em}{\includegraphics[width=0.95em]{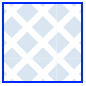}}}
\newcommand\allmark{\raisebox{-0.2em}{\includegraphics[width=1.1em]{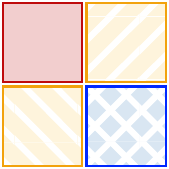}}}

\usepackage{wrapfig}

\newcommand{\figGraphicalAbstract}{
\setlength{\columnsep}{0.5em}
\setlength{\intextsep}{0.3em}
\begin{wrapfigure}{r}{0.5\textwidth}
\centering
\includegraphics[width=0.95\textwidth]{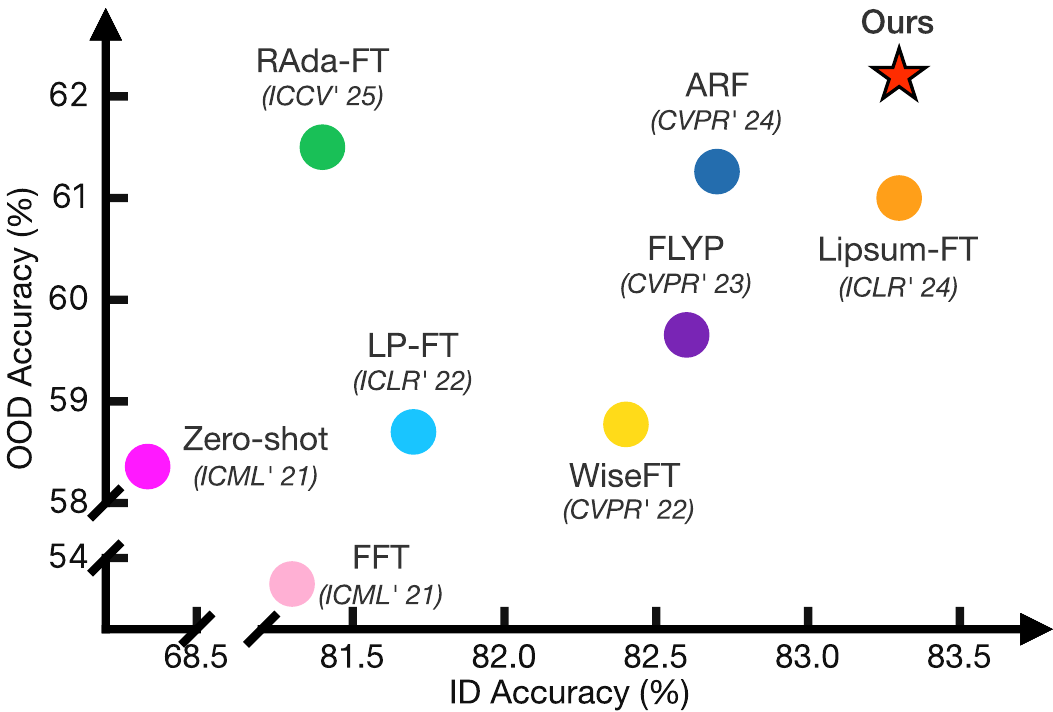}
\caption{\label{fig:graphical-abstract}
ID and OOD accuracy (\%) of state-of-the-art models fine-tuned on the ImageNet dataset using CLIP ViT-B/16. Ours (red star) achieves the best ID-OOD performance.
}
\end{wrapfigure}
}

\newcommand{\figTrainingProcess}{
\begin{figure*}[!t]
\centering
\includegraphics[width=0.98\textwidth]{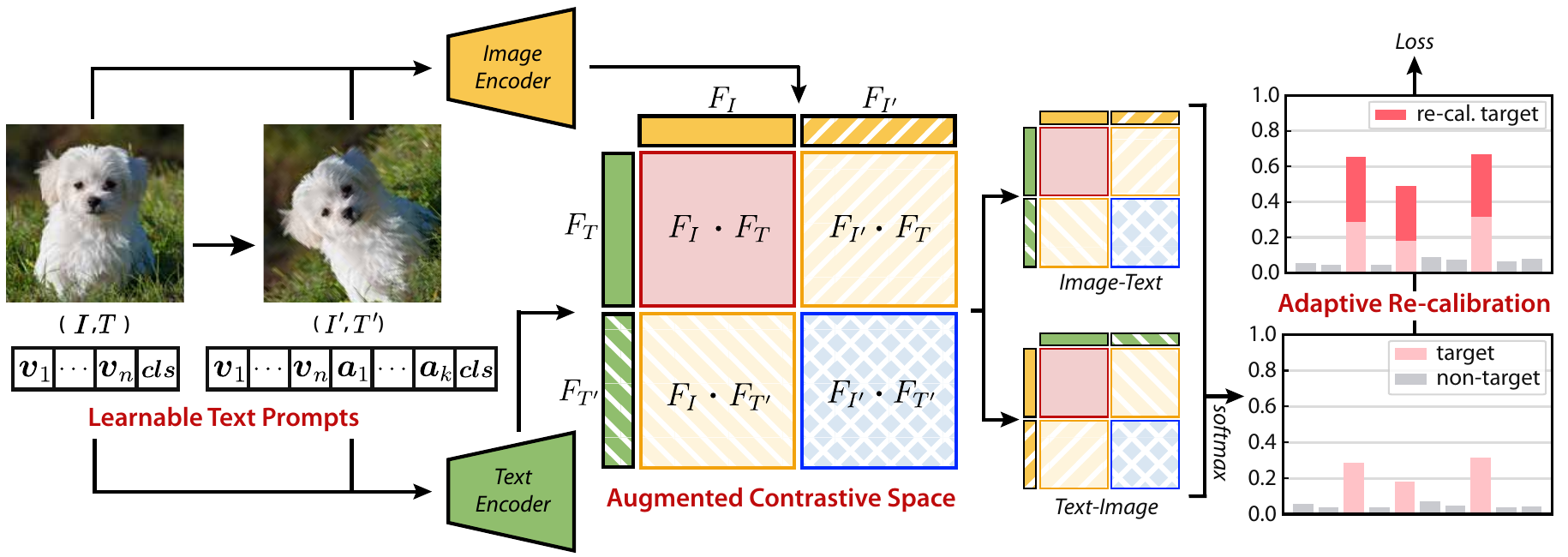}
\caption{\label{fig:training-process}
Overview of the proposed training scheme. Each clean image and prompt produces a transformed image and a transformation-conditioned prompt. The encoders extract four types of contrastive matches: clean features $(F_I, F_T)$ and augmented features $(F_{I'}, F_{T'})$. These features construct an augmented contrastive space. We apply a re-calibrated contrastive loss to update the networks. In learnable text prompts, $[v_1, v_2, \dots, v_n]$ and $[a_1, a_2, \dots, a_k]$ denote learnable context vectors and fixed transformation descriptor embeddings, respectively.
}
\end{figure*}
}

\newcommand{\figLoss}{
\begin{figure*}[!t]
        \centering
        \hspace{1.5em}
        \begin{minipage}{0.5\linewidth}
        \centering
        \includegraphics[width=0.9\textwidth]{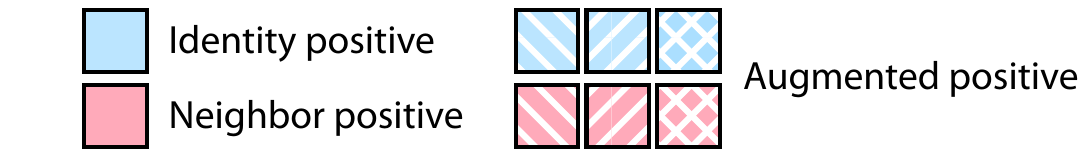}
        \subcaptionbox{CLIP\label{fig:loss-clip}}{\includegraphics[width=0.38\textwidth]{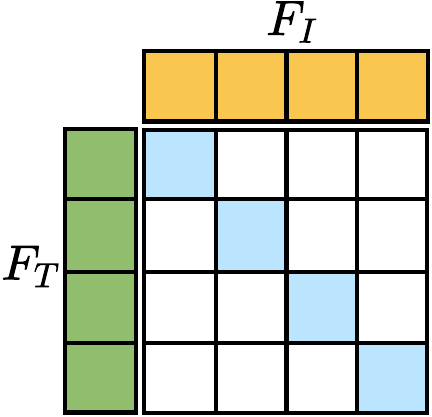}}
        \hspace{1.5em}
        \subcaptionbox{CLIP w/ SoftCE\label{fig:loss-softce}}{\includegraphics[width=0.38\textwidth]{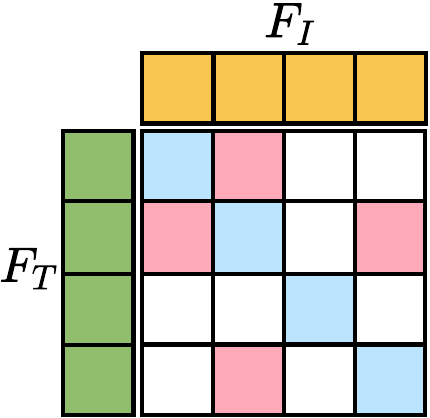}}
        \end{minipage}
        \hspace{-3.0em}
        \begin{minipage}{0.45\linewidth}
        \centering
        \subcaptionbox{Ours\label{fig:loss-ours}}{\includegraphics[width=0.6\textwidth]{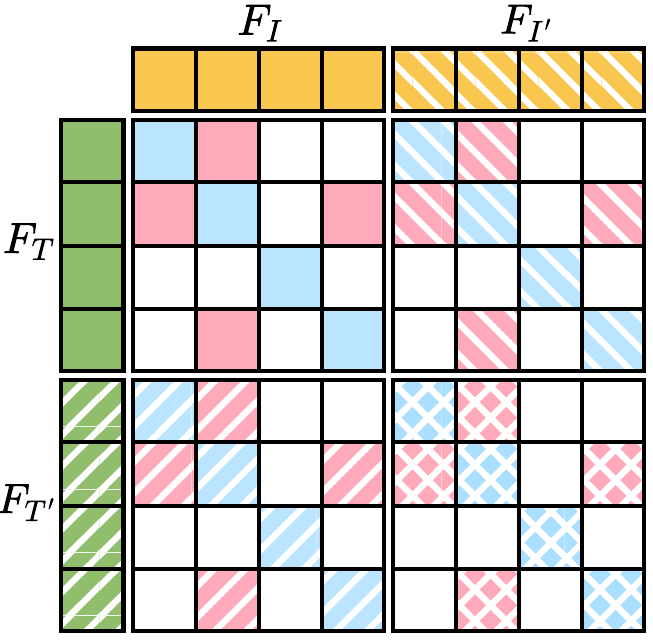}}
        \end{minipage}
\caption{\label{fig:loss}
Illustration of positive sampling used by different contrastive objectives. \textbf{(a)} CLIP uses only paired samples as positives. \textbf{(b)} CLIP with soft cross-entropy adds same-class positives. \textbf{(c)} Ours augments the positive samples in the contrastive space from $F_{I'}$ and $F_{T'}$.
}
\end{figure*}
}

\newcommand{\figSaturationEpochwise}{
\begin{figure*}[!t]
    \centering
    \subcaptionbox{Loss (FLYP)\label{fig:saturation-flyp}}{\includegraphics[width=0.24\textwidth]{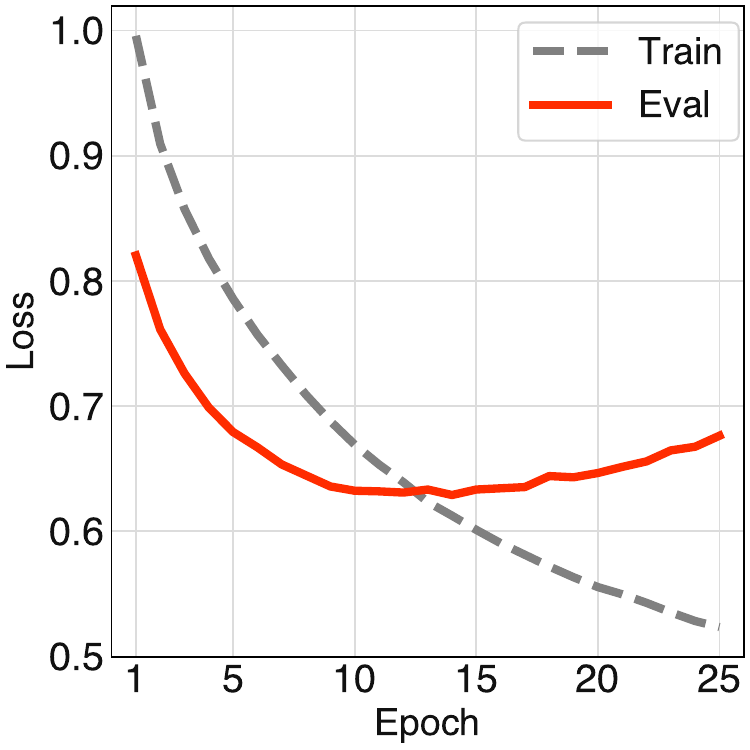}}
    \subcaptionbox{Loss (Ours)\label{fig:saturation-ours}}{\includegraphics[width=0.24\textwidth]{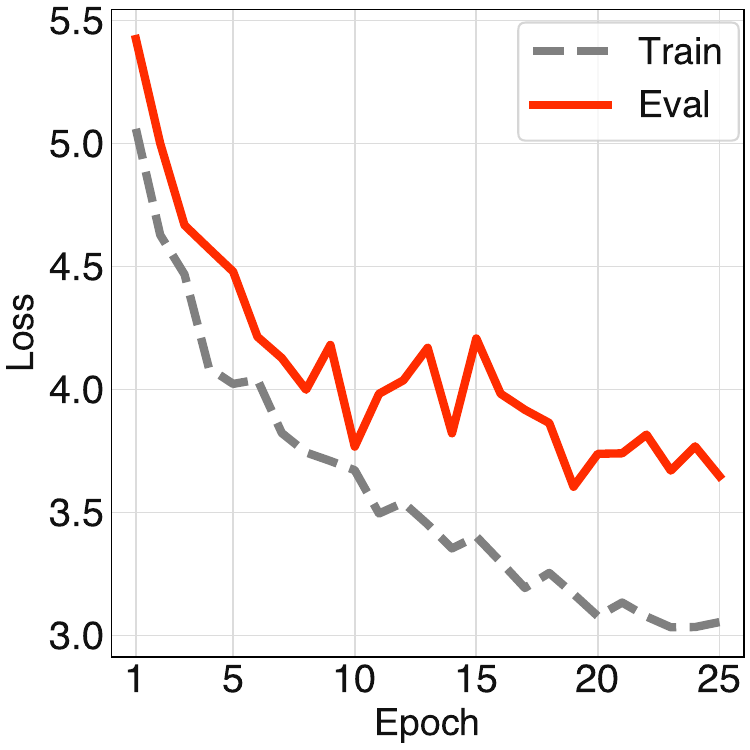}}
    \subcaptionbox{Accuracy (ID)\label{fig:epochwise-id}}{\includegraphics[width=0.24\textwidth]{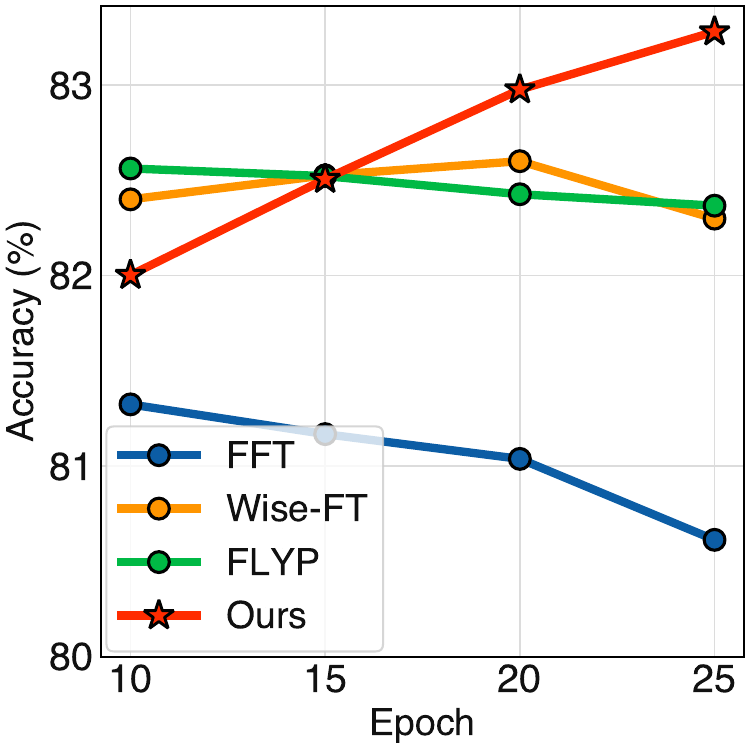}}
    \subcaptionbox{Accuracy (H.M.)\label{fig:epochwise-hmean}}{\includegraphics[width=0.24\textwidth]{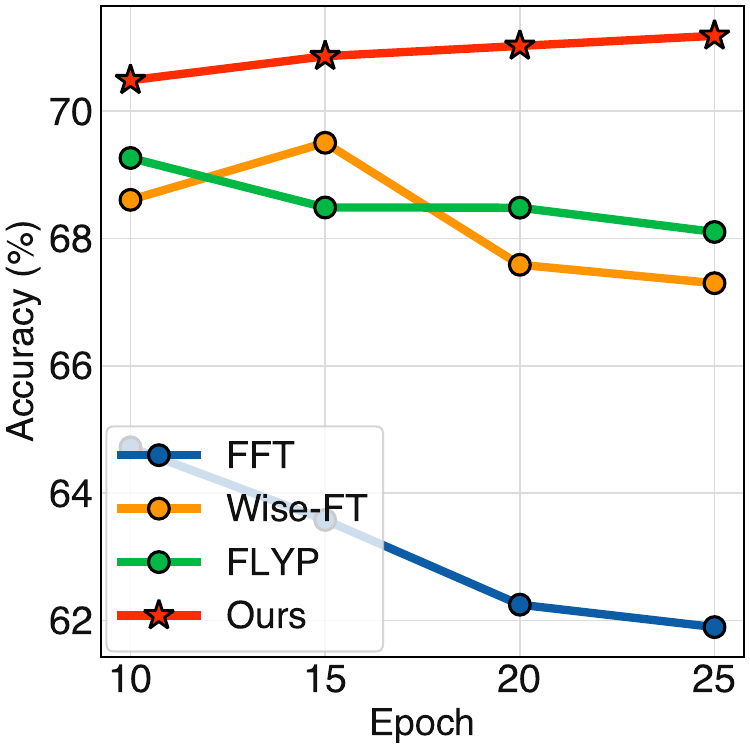}}
\caption{\label{fig:saturation-epochwise}
\textbf{(a-b)} Training and evaluation losses for ImageNet transfer learning with ViT-B/16.
\textbf{(c-d)} ID accuracy and harmonic mean under extended training.
}
\end{figure*}
}

\newcommand{\figFewShot}{
\begin{figure*}[!t]
\centering
\includegraphics[width=0.95\linewidth]{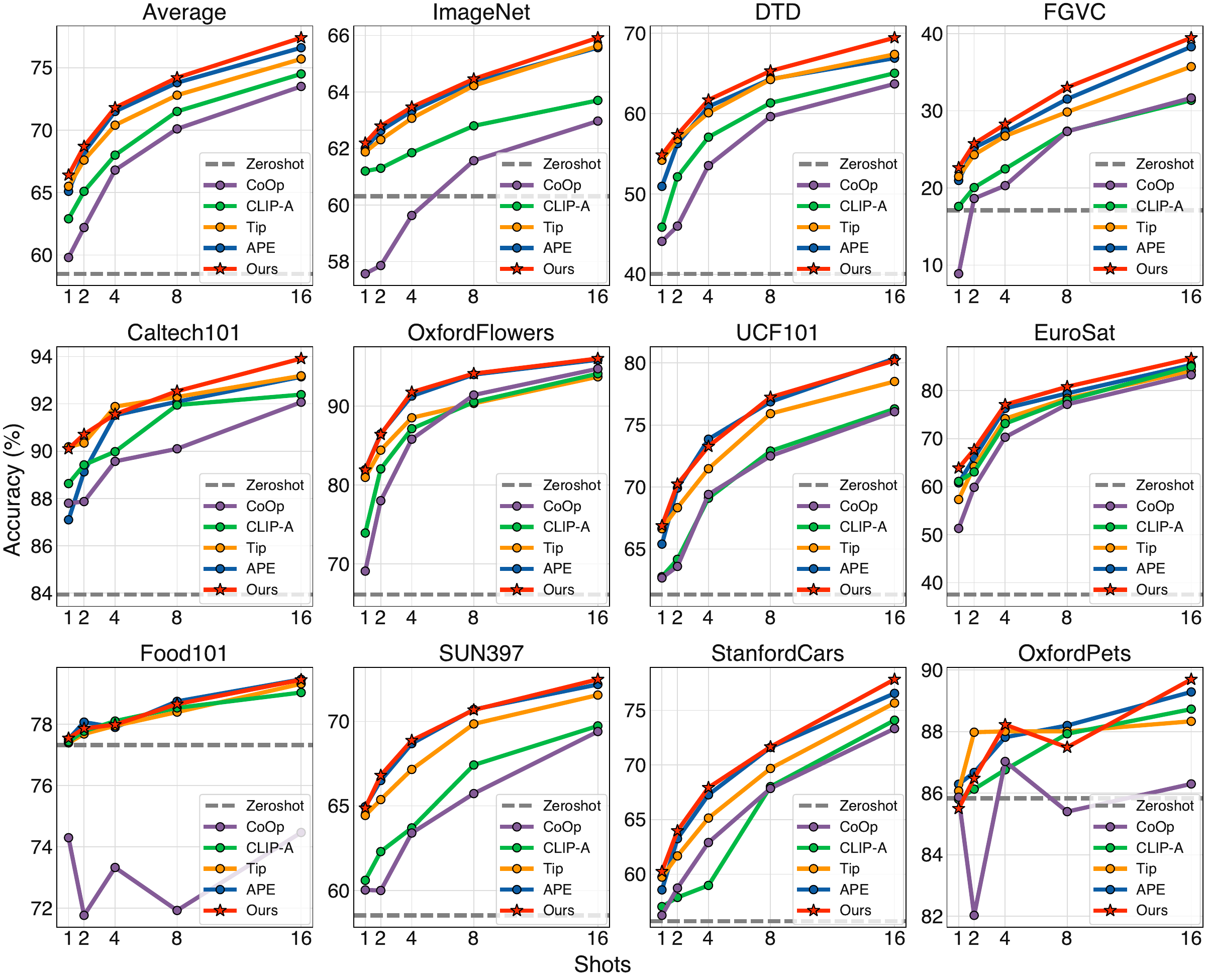}
\caption{\label{fig:fewshot}
Experimental results of few-shot learning using a ResNet50 backbone. We compare training-based adaptation methods with zero-shot CLIP on 11 few-shot benchmarks.
}
\end{figure*}
}

\newcommand{\SupFigEpochwise}{
\begin{figure*}[!hb]
\centering
\includegraphics[width=0.9\linewidth]{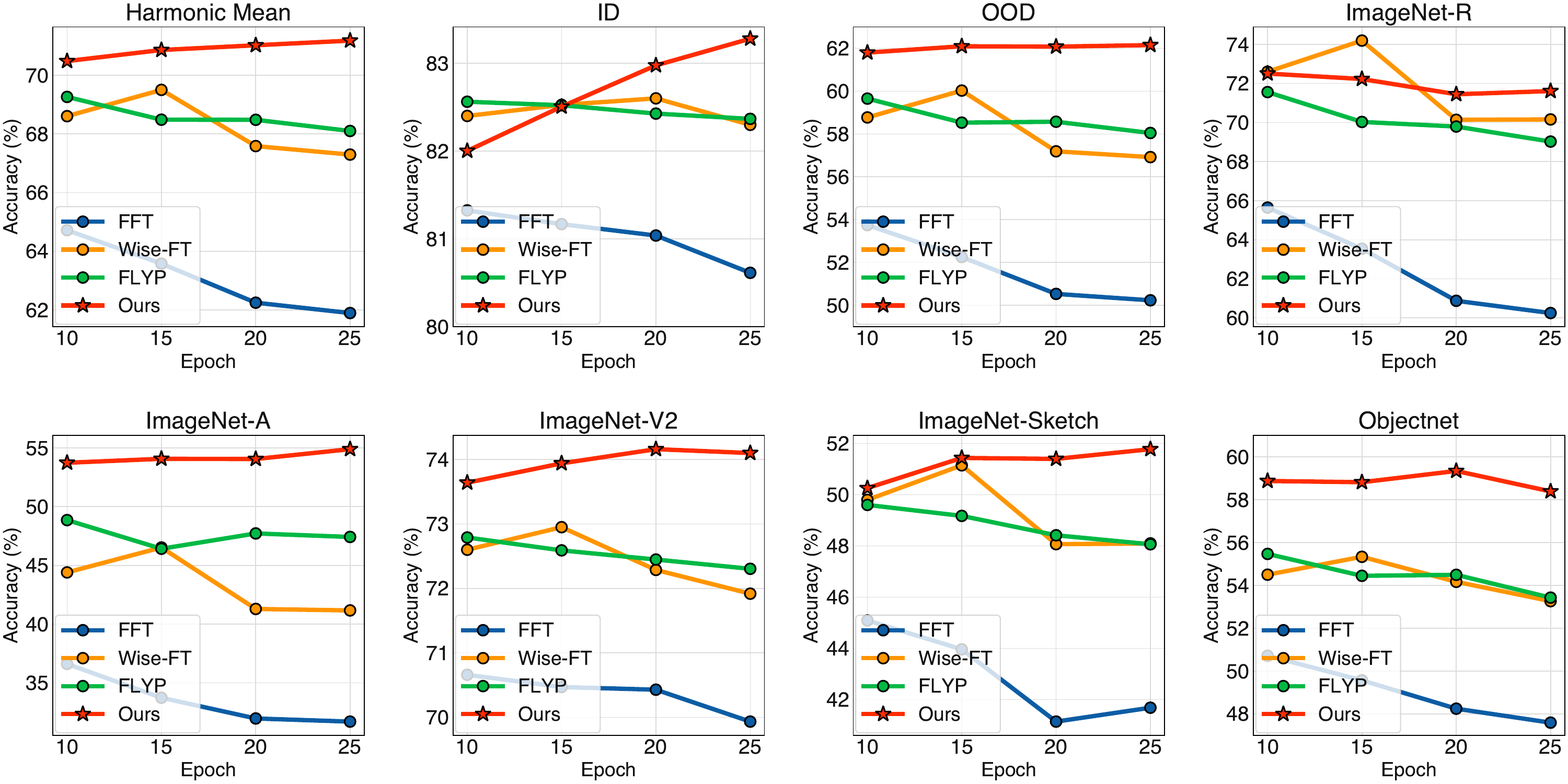}
\caption{\label{fig:sup-epochwise}
Additional experimental results of extended training with ViT-B/16.}
\end{figure*}
}

\usepackage{multirow, graphicx, colortbl, floatrow}
\newcommand{\CC}{\cellcolor{lightgray!30}}

\newcommand{\tabDistributionShift}{
\begin{table*}[!t]
\small\centering\setlength{\tabcolsep}{2.5pt}
\begin{tabular}{cccccccccccccc}
\toprule
\multirow{3}{*}{Backbone} & \multirow{3}{*}{Archi.} & \multirow{3}{*}{Model} & \multicolumn{8}{c}{ImageNet} & \multicolumn{3}{c}{iWildCam} \\
& & & ID & R & A & V2 & Ske. & Obj. & O.M. & H.M. & ID & OOD & H.M.\\
\cmidrule{1-3} \cmidrule(l){4-11} \cmidrule(l){12-14}
\multirow{12}{*}{Update} & \multirow{8}{*}{B/16} & FFT & 81.3 & 65.6 & 36.6 & 70.7 & 45.1 & 50.7 & 53.7 & 64.7 & 50.3 & 34.1 & 40.6 \\
 &  & LP-FT & 81.7 & 73.5 & 42.5 & 72.1 & 50.3 & 55.1 & 58.7 & 68.3 & 49.7 & 34.7 & 40.9 \\
 &  & WiseFT & 82.4 & 72.6 & 44.4 & 72.6 & 49.8 & 54.5 & 58.8 & 68.6 & 48.5 & 37.2 & 42.1 \\
 &  & FLYP & 82.6 & 71.6 & 48.9 & 72.8 & 49.6 & 55.5 & 59.7 & 69.3 & 50.1 & 36.3 & 42.1 \\
 &  & ARF & 82.7 & 75.6 & 50.3 & 72.8 & 51.8 & 55.8 & 61.3 & 70.4 & \multicolumn{3}{c}{\multirow{1}{*}{-}} \\
 &  & Lipsum-FT & 83.3 & \textbf{75.9} & 49.9 & 73.6 & 51.4 & 54.4 & 61.0 & 70.5 & \multicolumn{3}{c}{\multirow{1}{*}{-}} \\
 &  & RAda-FT & 81.4 & 75.5 & 51.7 & 71.9 & 50.4 & 56.8 & 61.3 & 69.9 & \multicolumn{3}{c}{\multirow{1}{*}{-}} \\
 &  & \CC Ours & \CC \textbf{83.3} & \CC 71.6 & \CC \textbf{54.9} & \CC \textbf{74.1} & \CC \textbf{51.8} & \CC \textbf{58.4} & \CC \textbf{62.2} & \CC \textbf{71.2} & \CC \textbf{51.3} & \CC \textbf{37.8} & \CC \textbf{43.5} \\
 \cmidrule{2-3} \cmidrule(l){4-11} \cmidrule(l){12-14}
 & \multirow{4}{*}{B/32} & FFT & 76.2 & 57.2 & 20.3 & 64.3 & 39.5 & 40.6 & 44.4 & 56.1 & 37.2 & 22.6 & 28.1 \\
 &  & WiseFT & 77.6 & 65.4 & 28.2 & 66.9 & 45.4 & 45.2 & 50.2 & 61.0 & 40.8 & 27.0 & 32.5 \\
 &  & FLYP & 77.6 & \textbf{66.4} & 28.2 & 66.9 & 45.4 & 45.2 & 50.4 & 61.1 & 39.3 & 25.5 & 30.9 \\
 &  & \CC Ours & \CC \textbf{79.4} & \CC 62.4 & \CC \textbf{32.6} & \CC \textbf{68.2} & \CC \textbf{44.5} & \CC \textbf{49.2} & \CC \textbf{51.4} & \CC \textbf{62.4} & \CC \textbf{42.1} & \CC \textbf{29.3} & \CC \textbf{34.5} \\
 \cmidrule{1-3} \cmidrule(l){4-11} \cmidrule(l){12-14}
\multirow{5}{*}{Freeze} & \multirow{3}{*}{B/16} & Zero-shot & 68.3 & \textbf{77.7} & 49.9 & 61.9 & 48.3 & 54.0 & 58.4 & 63.0 & 8.7 & 11.0 & 9.7 \\
 &  & LinearProb & 79.9 & 70.8 & 46.4 & 69.8 & 46.9 & 52.1 & 57.2 & 66.7 & 44.5 & \textbf{31.1} & \textbf{36.6} \\
 &  & \CC Ours (PETL) & \CC \textbf{80.1} & \CC 73.5 & \CC \textbf{51.4} & \CC \textbf{70.3} & \CC \textbf{49.4} & \CC \textbf{55.7} & \CC \textbf{60.1} & \CC \textbf{68.7} & \CC \textbf{45.5} & \CC 30.3 & \CC 36.4 \\
 \cmidrule{2-3} \cmidrule(l){4-11} \cmidrule(l){12-14}
 & \multirow{2}{*}{B/32} & Zero-shot & 63.4 & 69.3 & 31.6 & 56.0 & 42.3 & 44.4 & 48.7 & 55.1 & 5.3 & 7.3 & 6.1 \\
 &  & \CC Ours (PETL) & \CC 76.1 & \CC 63.6 & \CC 30.5 & \CC 64.5 & \CC 42.3 & \CC 46.7 & \CC 49.5 & \CC 60.0 & \CC 34.7 & \CC 24.5 & \CC 28.7 \\
\bottomrule
\end{tabular}
\caption{Experimental results of the distribution shift on ImageNet and iWildCam. We compare Ours with previous methods of backbone update and freeze training.
The `O.M.' and `H.M.' denote the OOD mean and harmonic mean of ID and OOD mean.
}
\label{tab:distribution-shift}
\end{table*}
}

\newcommand{\tabTransferLearning}{
\begin{table*}[!t]
\small\centering\setlength{\tabcolsep}{2.5pt}
\begin{tabular}{cccccccc}
\toprule
Model & Caltech101 & PCam & ImageNet & Flowers102 & StanfordCars & iWildCam & Total Mean \\
\midrule
Zero-shot & 88.7 & 54.0 & 68.3 & 71.0 & 64.7 & 8.7 & 59.3 \\
LinearProb & 94.8 & 82.6 & 79.9 & 95.9 & 83.1 & 44.5 & 80.1 \\
FFT & 97.2 & 89.1 & 81.4 & 90.4 & 84.4 & 50.3 & 82.1 \\
LP-FT & 96.9 & 89.0 & 81.8 & 97.9 & 89.4 & 49.7 & 84.1 \\
FLYP & 97.6 & \textbf{90.3} & 82.6 & 97.7 & 89.6 & 50.1 & 84.6 \\
\rowcolor{lightgray!30} Ours & \textbf{97.9} & 89.4 & \textbf{83.3} & \textbf{98.9} & \textbf{91.4} & \textbf{51.3} & \textbf{85.4} \\
\bottomrule
\end{tabular}
\caption{Experimental results of transfer learning on six downstream datasets. We use ViT-B/16 as a backbone network for all methods.}
\label{tab:transfer-learning}
\end{table*}
}

\newcommand{\tabAugmentationAlignment}{
\begin{table*}[!t]
\small\centering\setlength{\tabcolsep}{2.5pt}
\begin{tabular}{cccccccccc}
\toprule
Image & Text & ID & R & A & V2 & Ske. & Obj. & OOD Mean & H.M. \\
\midrule
\multicolumn{2}{c}{CLIP} & 63.4 & 69.3 & 31.6 & 56.0 & 42.3 & 44.4 & 48.7 & 55.1 \\
Random Crop & Fix & 78.0 & 59.6 & 29.8 & 66.0 & 41.5 & 47.6 & 48.9 & 60.1 \\
Random Crop & Misalign. & 71.2 & 56.6 & 27.1 & 60.5 & 38.0 & 43.4 & 45.1 & 55.2 \\
RandAug & Fix & 78.6 & 61.7 & 31.8 & 68.2 & 43.5 & 48.3 & 50.7 & 61.6 \\
RandAug & Misalign. & 73.1 & 58.4 & 29.4 & 63.0 & 40.3 & 44.9 & 47.2 & 57.3 \\
\rowcolor{lightgray!30} RandAug & Align. & 79.4 & 62.4 & 32.6 & 68.2 & 44.5 & 49.2 & 51.4 & 62.4 \\
\bottomrule
\end{tabular}
\caption{Experimental results of transformation-aware prompt conditioning on ViT-B/32 backbone. A misaligned fixed descriptor consistently degrades performance, whereas the correct transformation descriptor improves performance.}
\label{tab:augmentation-alignment}
\end{table*}
}

\newcommand{\tabAblationStudy}{
\begin{table*}[!t]
\centering\small\setlength{\tabcolsep}{2.5pt}
\begin{tabular}{ccccccccccc}
\toprule
Adaptive Re-cal. & Space & Combination & ID & R & A & V2 & Ske. & Obj. & OOD Mean & H.M. \\
\midrule
- & \itmark & I$\leftrightarrow$T & 77.2 & 61.3 & 31.3 & 66.1 & 42.1 & 47.1 & 49.6 & 60.4 \\
- & \iptpmark & \Ip$\leftrightarrow$\Tp & 77.4 & 63.1 & 31.7 & 66.8 & 43.3 & 48.2 & 50.6 & 61.2 \\
- & \itmark+\iptpmark & I$\leftrightarrow$T, \Ip$\leftrightarrow$\Tp & 78.1 & 63.0 & 32.0 & 67.4 & 43.4 & 48.4 & 50.8 & 61.6 \\
- & \allmark & [I \Ip]$\leftrightarrow$[T \Tp] & 78.2 & 63.2 & 32.0 & 67.3 & 43.5 & 48.4 & 50.9 & 61.6 \\
\midrule
\checkmark & \itmark & I$\leftrightarrow$T & 78.5 & 61.9 & 32.3 & 68.0 & 43.5 & 48.0 & 50.7 & 61.6 \\
\checkmark & \iptpmark & \Ip$\leftrightarrow$\Tp & 78.4 & 63.0 & 32.4 & 67.3 & 43.9 & 48.3 & 51.0 & 61.8 \\
\checkmark & \itmark+\iptpmark & I$\leftrightarrow$T, \Ip$\leftrightarrow$\Tp & 79.0 & 62.7 & 32.6 & 68.0 & 43.8 & 48.4 & 51.1 & 62.0 \\
\rowcolor{lightgray!30} \checkmark & \allmark & [I \Ip]$\leftrightarrow$[T \Tp] & 79.4 & 62.4 & 32.6 & 68.2 & 44.5 & 49.2 & 51.4 & 62.4 \\
\bottomrule
\end{tabular}
\caption{Experimental results of our re-calibrated contrastive loss and contrastive space on ViT-B/32 backbone. Performance is measured in terms of accuracy with respect to the applied loss function. I$\leftrightarrow$T denotes image-text and text-image match (\eg, \itmark \hquad red region in \cref{fig:training-process}) and \Ip$\leftrightarrow$\Tp{} is a transformation-conditioned augmented match (\eg, \iptpmark \hquad blue region in \cref{fig:training-process}). [I \Ip]$\leftrightarrow$[T \Tp] represents the four contrastive matches of all clean and transformation-conditioned augmented image-text pairs (\eg, \allmark \hquad whole region of the augmented contrastive space).}
\label{tab:ablation-study}
\end{table*}
}

\newcommand{\tabObjectiveComparison}{
\begin{table*}[!t]
\small\centering\setlength{\tabcolsep}{2.5pt}
\begin{tabular}{lcccccccc}
\toprule
Objective / Configuration & ID & R & A & V2 & Ske. & Obj. & OOD Mean & H.M. \\
\midrule
InfoNCE & 73.7 & 56.3 & 26.1 & 63.1 & 40.4 & 46.1 & 46.4 & 57.0 \\
Focal & 78.5 & \textbf{63.3} & \textbf{32.8} & 67.5 & 43.3 & 48.3 & 51.0 & 61.8 \\
SoftCE (=SupCon) & 78.4 & 63.1 & 32.4 & 67.6 & 43.4 & 48.4 & 51.0 & 61.8 \\
\midrule
$\enspace+$ Re-calibrated & 79.0 & 62.6 & 32.6 & 67.8 & 44.1 & 48.6 & 51.1 & 62.1 \\
$\;\;\enspace+$ Multi-sample fusion & \textbf{79.4} & 62.4 & 32.6 & \textbf{68.2} & \textbf{44.5} & \textbf{49.2} & \textbf{51.4} & \textbf{62.4} \\
\bottomrule
\end{tabular}
\caption{Objective and component comparison on ImageNet distribution using ViT-B/32.
All configurations use transformation-aware prompt conditioning and the
augmented contrastive space. The last two rows incrementally add
re-calibration and multi-sample fusion to SoftCE.}
\label{tab:objective-comparison}
\end{table*}
}

\newcommand{\SupTabHyperparameter}{
\begin{table*}[!ht]
\thisfloatsetup{capposition=bottom}
\begin{floatrow}
\ttabbox{
\setlength{\tabcolsep}{10pt}
\begin{tabular}{cc}
\toprule
Hyperparameter & Setting \\ \midrule
Optimizer & AdamW \\
Betas & (0.9, 0.999) \\
Scheduler & cosine \\
Warmup LR / epoch & 0.0 / 1\\
RandAug & (7, 0.5) \\
Interpolation & bicubic \\
\bottomrule
\end{tabular}
}
{\caption{Default hyperparameters used in our experiments. LR denotes learning rate.}
\label{tab:sup-hyperparameter-default}}

\ttabbox{
\setlength{\tabcolsep}{15pt}
\begin{tabular}{cc}
\toprule
Hyperparameter & Setting \\
\midrule
Backbone & B/16 \\
Batch size & 640 \\
Epoch & 100 \\
LR & $5.0e^{-6}$\\
Adapter LR & $4.0e^{-4}$\\
Weight decay & $1.0e^{-2}$\\
\bottomrule
\end{tabular}}
{\caption{Hyperparameters of the transfer learning task.}
\label{tab:sup-hyperparameter-transfer}}
\end{floatrow}
\end{table*}
}

\newcommand{\SupTabHyperparameterImageNet}{
\begin{table}[!ht]
\centering
\begin{tabular}{ccccc}
\toprule
Backbone & B/16 & B/16$^\dag$ & B/32 & B/32$^\dag$ \\
\midrule
Batch size & 640 & 2048 & 512 & 2048 \\
Epoch & 25 & 25 & 25 & 25 \\
LR & $1.25e^{-6}$ & - & $1.25e^{-6}$ & - \\
Adapter LR & $1.0e^{-4}$ & $1.0e^{-3}$ & $1.0e^{-4}$ & $1.0e^{-3}$ \\
Weight decay & $1.0e^{-2}$ & $1.0e^{-1}$ & $1.0e^{-2}$ & $1.0e^{-2}$ \\
\bottomrule
\end{tabular}
\caption{Details of hyperparameter settings for the distribution shift experiments on ImageNet.}
\label{tab:sup-hyperparameter-imagenet}
\end{table}
}

\newcommand{\SupTabHyperparameterIWildCam}{
\begin{table}[!ht]
\centering
\begin{tabular}{ccccc}
\toprule
Backbone & B/16 & B/16$^\dag$ & B/32 & B/32$^\dag$ \\
\midrule
Batch size & 640 & 2048 & 512 & 2048 \\
Epoch & 45 & 25 & 45 & 25 \\
LR & $5.0e^{-6}$ & - & $5.0e^{-6}$ & - \\
Adapter LR & $1.0e^{-4}$ & $1.0e^{-2}$ & $1.0e^{-4}$ & $1.0e^{-2}$ \\
Weight decay & $1.0e^{-2}$ & $1.0e^{-2}$ & $1.0e^{-2}$ & $1.0e^{-2}$ \\
\bottomrule
\end{tabular}
\caption{Details of hyperparameter settings for the distribution shift experiments on iWildCam.}
\label{tab:sup-hyperparameter-iwildcam}
\end{table}
}

\newcommand{\SupTabGradientDiagnostics}{
\begin{table*}[!ht]
\small\centering\setlength{\tabcolsep}{5pt}
\begin{tabular}{lccccc}
\toprule
$|\bm{P}(i)|$ & 1 & 2 & 4 & 8 & 16 \\
\midrule
Gradient ratio (ours / SoftCE) & 1.46 & 1.68 & 1.79 & 1.89 & 1.81 \\
\bottomrule
\end{tabular}
\caption{Gradient-norm ratio between the re-calibrated objective and SoftCE.}
\label{tab:sup-gradient-diagnostics}
\end{table*}
}

\newcommand{\SupTabEfficiency}{
\begin{table}[!ht]
\small\centering\setlength{\tabcolsep}{4pt}
\begin{tabular}{lcccc}
\toprule
Batch size $B$ & 8 & 16 & 32 & 64 \\
\midrule
Runtime (ours / FLYP) & $1.0\times$ & $1.1\times$ & $1.6\times$ & $1.7\times$ \\
Peak memory (ours / FLYP) & $1.0\times$ & $1.2\times$ & $1.4\times$ & $1.6\times$ \\
\bottomrule
\end{tabular}
\caption{Per-step training overhead relative to FLYP.}
\label{tab:sup-efficiency}
\end{table}
}

\newcommand{\SupTabTransformationCount}{
\begin{table*}[!ht]
\small\centering\setlength{\tabcolsep}{3pt}
\begin{tabular}{lcccccccc}
\toprule
$k$ & ID & R & A & V2 & Ske. & Obj. & OOD Mean & H.M. \\
\midrule
1 & 75.8 & 61.6 & 31.5 & 64.9 & 41.9 & 48.7 & 49.7 & 60.0 \\
2 (default) & 79.4 & 62.4 & 32.6 & 68.2 & 44.5 & 49.2 & 51.4 & 62.4 \\
3 & 79.4 & 63.2 & 32.8 & 68.6 & 44.6 & 49.3 & 51.7 & 62.6 \\
\bottomrule
\end{tabular}
\caption{Sensitivity to the number of sampled image transformations $k$.}
\label{tab:sup-transformation-count}
\end{table*}
}

\title{Re-calibrated Contrastive Loss for \\ Transformation-Aware Prompt Conditioning in Vision-Language Models}

\addauthor{Seungmin Oh}{seungminoh@ajou.ac.kr}{1}
\addauthor{Seunghun Kang}{seunghun.kang@ajou.ac.kr}{1}
\addauthor{Jongbin Ryu$^\ast$}{jongbinryu@ajou.ac.kr}{1,2}

\addinstitution{
 Department of Artificial Intelligence,\\Ajou University,\\Suwon, South Korea
}
\addinstitution{
 Department of Computer Engineering,\\Ajou University,\\Suwon, South Korea
}

\runninghead{Oh, Kang, Ryu}{Re-calibrated Contrastive Loss for Prompt Conditioning}

\def\eg{\emph{e.g}\bmvaOneDot}

\begin{document}

\maketitle

\begin{abstract}
Ensuring effective transfer learning for vision-language models without compromising their generalization performance is crucial. However, many existing methods overlook data characteristics and simply reuse the training strategies adopted during pre-training. Specifically, they treat same-class samples as distinct instances and transform images independently of their paired text prompts, which makes model learning more difficult. We address these limitations through transformation-aware prompt conditioning and a re-calibrated contrastive loss. Fixed text descriptors identify the transformations applied to paired images, providing transformation-level consistency without altering class semantics. This design aligns the image and text branches at the transformation level, enabling richer representations while preserving the models' ability to generalize. In addition, our loss function mitigates positive-gradient dilution in soft-target cross-entropy when each anchor has multiple valid positives. During transfer, our approach treats same-class samples as positives rather than distinct instances, enabling the model to learn domain-specific features more effectively. Experiments across distribution shift, transfer learning, and few-shot settings demonstrate consistent improvements over existing approaches. Source code for our method is available at \url{https://github.com/SoongE/ReCalCon}.
\end{abstract}
    
\section{Introduction}
\label{sec:introduce}
Interest in Vision-Language Models (VLMs) has increased due to their capacity for generalized vision-language representation.
This capability enables efficient adaptation to target domains through fine-tuning
~\citep{wortsman2022wiseft, goyal2023flyp, kumar2022lpft, gao2023clipadapter, zhang2021tip, zhou2022coop, zhu2023ape}.
However, it is important to point out that the current transfer learning approach presents two challenges: 1) overfitting to the target domain and 2) a discrepancy between augmented images and their text prompts.
The first problem arises when fine-tuning the pre-trained model to a specific domain, as this may result in subpar performance in other domains due to distribution shifts, as shown in \cref{fig:graphical-abstract}.

\figGraphicalAbstract
Full fine-tuning (FFT) of zero-shot models can create a trade-off between in-distribution (ID) and out-of-distribution (OOD) accuracy.
For several transfer baselines, this trade-off becomes more pronounced as training continues. 
The second challenge arises because text prompts often do not describe image transformations.
Most models directly adopt the contrastive learning approach used during pre-training, which is typically based on a single image-text positive pair. As a result, while images undergo various augmentations, text prompts either remain fixed, rely on generative models, or include meaningless random words. Consequently, visual data is typically trained using extensive augmentation techniques, whereas text prompts fail to reflect these augmentations accurately.

We address this discrepancy by conditioning each text prompt on the transformation applied to its paired image.
We call the method \textbf{Transformation-Aware Prompt Conditioning}.
Transformation-level consistency pairs each augmented image with a fixed descriptor of the applied transformation without implying a change in class semantics.
As shown in \cref{fig:training-process}, we expand the contrastive space to include augmented images and transformation-conditioned prompts. The expanded space contains four types of contrastive matches.
While only the contrastive match between clean images and prompts (red in \cref{fig:training-process}) has been used in previous VLM transfer learning approaches, we exploit additional matches in the augmented space (blue in \cref{fig:training-process}).
We also include cross-pairs between the clean and augmented data (yellow in \cref{fig:training-process}).
These additional matches provide regularization during VLM transfer.

Expanding the contrastive space introduces multiple valid positives for each anchor, whereas the pre-training objective uses only one paired positive.
A uniform soft target distributes the target probability mass equally across these positives. The learning signal assigned to each positive therefore decreases as the number of positives increases. We refer to this effect as positive-gradient dilution.
To address this, we propose a \textbf{Re-calibrated Contrastive Loss} function that adds an adaptive positive-versus-negative evidence margin while leaving the normalized softmax distribution unchanged. Our loss function effectively accounts for multiple positive pairs within a single logit vector, enabling the learning of both symmetric and asymmetric contrastive matches.
Together, the re-calibrated contrastive loss and transformation-aware prompt conditioning transfer VLMs to target domains, demonstrating strong performance on out-of-distribution data. 
We also show that our approach performs much better, as the networks are trained longer without overfitting the target training data.
This is because our approach to VLM transfer learning effectively regularizes, leading to performance improvements on out-of-distribution data, even with longer training.
We perform various experiments evaluating transfer learning using ImageNet~\citep{deng2009imagenet} and iWildCam~\citep{beery2020iwildcam} datasets for in- and out-of-distribution data classification. 
We also provide experimental results for transfer learning and few-shot scenarios across diverse downstream datasets.

\figTrainingProcess
\subsection{Contributions}
\label{sec:contributions}
\paragraph{Generic method.} 
We combine the re-calibrated contrastive loss with transformation-aware prompt conditioning for VLM transfer learning.
The framework jointly optimizes clean, transformed, and cross-paired image-text matches in a transformation-consistent space.
Our experiments show that the proposed approach improves performance across various transfer learning tasks compared to the previous state-of-the-art methods.

\paragraph{Simplicity.}
The proposed transformation-aware prompt conditioning method follows a straightforward strategy, making it easy to integrate existing visual augmentation techniques into our approach. The re-calibrated contrastive loss requires only a few lines of code.

\paragraph{Regularization.}
Overfitting under distribution shift often becomes more severe as training continues. Several existing VLM transfer baselines therefore degrade under extended training, whereas our regularization allows performance to continue improving.

\figLoss

\section{Related Work}

\subsection{Transferring Vision-Language Models}
Pre-trained Vision-Language Models (VLMs) can be adapted to downstream tasks with limited target-domain data. Image classification is a common setting for this transfer.

Studies on image classification transfer learning can be categorized into prompt tuning~\citep{zhou2022coop, shu2022tpt, khattak2023maple}, network adapters~\citep{gao2023clipadapter, zhang2021tip, zhu2023ape}, and full fine-tuning~\citep{kumar2022lpft, wortsman2022wiseft, goyal2023flyp}.
Prompt tuning is often used for parameter-efficient transfer learning, while network adapters are primarily employed for few-shot learning when the model is transferred with limited training samples.
This VLM-based transfer learning strategy offers the advantage of efficient and fast adaptation to various domains by leveraging the general knowledge learned from image-text pairs.

\subsection{Distribution Shift}
The primary challenge in transferring pre-trained VLMs~\citep{radford2021clip, li2021albef, alayrac2022flamingo} is that the generalized feature distribution can shift toward a particular domain.
Therefore, it is critical to ensure that the distribution remains general while enabling model transferability.
WiseFT~\citep{wortsman2022wiseft} uses a weight ensemble method for shifted distributions after transfer learning to preserve the generalized distribution.
LP-FT~\citep{kumar2022lpft} first trains only a linear classifier and then fine-tunes the entire network to balance generalization and transferability.
Neither of these methods updates the text encoder during the transfer learning of the pre-trained VLM.
Lipsum-FT~\citep{nam2024lipsumft} adds random text while freezing the text encoder, and RAda-FT~\citep{chen2025radaft} applies a learnable attention mask to the final fused representation.
In contrast, FLYP~\citep{goyal2023flyp} updates both encoders using a CLIP-style contrastive objective on image-text pairs. ARF~\citep{han2024arf} enriches text prompts with context generated by large language models while updating the network.

\subsection{Parameter-Efficient Transfer Learning}
Fine-tuning all layers of a network involves a large number of trainable parameters. However, numerous studies~\citep{donahue2014decaf, grill2020byol, yosinski2014transferable} have shown that models pre-trained on large-scale datasets can be adapted without fine-tuning every layer. Adapters~\citep{gao2023clipadapter, yu2023taskresidual} introduce small, learnable modules that capture task-specific features while leveraging frozen pre-trained weights. By combining features with newly learned adapter features via a residual connection, these approaches maintain VLMs' zero-shot capability while enhancing performance in few-shot scenarios.
Prompt tuning~\citep{zhou2022coop, shu2022tpt} replaces manually created prompts with small learnable vectors. These vectors are refined via backpropagation on a limited number of samples, enabling the model to learn task-specific descriptions without manual prompt design. 
Prior-based methods~\citep{zhang2021tip, zhu2023ape} capitalize on the inherent data distribution captured by the pre-trained model, rather than introducing new parameters. These methods create a cache model from few-shot training samples, decoupling the specific knowledge from the general cache.

\subsection{Advances in Cross-Entropy Loss}
Several studies have addressed the limitations associated with using cross-entropy loss with the softmax function.
To address imbalanced class distributions, \citet{lin2017focalloss} proposed assigning higher weights to positive samples while reducing the weights of negative samples.
\citet{chen2017noisysoftmax} introduced a noise-based regularization method to ensure continuous gradient propagation and prevent early-saturated loss values caused by the softmax function.
\citet{khosla2020supervised} extended contrastive objectives to multiple same-class positives.

\section{Method}
This section introduces transformation-aware prompt conditioning for image and text prompts in a shared space, followed by the introduction of the proposed re-calibrated contrastive loss.

\subsection{Transformation-Aware Prompt Conditioning}
A proper strategy must account for the differences between image and text data when augmenting them in the shared space.
Recent studies have demonstrated that the performance of vision-language models varies with their prompt design~\citep{radford2021clip, zhang2021tip, an2024perceptionclip}.
Therefore, we introduce a learnable text prompt to efficiently describe the image augmentation in a text encoder, as shown in \cref{fig:training-process}. 
We construct the learnable text prompt $\mathcal{M}$ by concatenating the learnable context $V$, fixed transformation descriptor $A$, and class token $\text{CLS}$ as:
\begin{gather*}
\mathcal{M} = [V, A,\text{CLS}], \quad
V = [v_1, v_2, \dots, v_n]^\top \in \mathbb{R}^{n \times m},\\
A = [a_1, a_2, \dots, a_k]^\top \in \mathbb{R}^{k \times m},\quad
\text{CLS} = [cls] \in \mathbb{R}^{1 \times m}.
\end{gather*}
$\mathcal{M}$ differs from \citet{zhou2022coop} in that it includes $A$, which describes the augmentation applied to the image.
$A$ is a fixed, non-learnable textual descriptor from the RandAug~\citep{cubuk2020randaug} dictionary, whereas $V$ is the learnable CoOp~\citep{zhou2022coop} context. Thus, only $V$ is optimized and $A$ is never updated.
We use this learnable text prompt as input to a text encoder, allowing the learnable context vector to be optimized during training. Additionally, we develop a pool-based strategy for applying sophisticated image augmentation methods. A pool of augmentation methods is created for training deep neural networks, from which we randomly select $k$ methods to apply combined augmentations to an image. $\text{CLS}$ is the word embedding vector from the text encoder. 
We set $n=4$ following~\citet{zhou2022coop}, $m=512$ to match the encoder dimension, and $k=2$ following~\citet{cubuk2020randaug}.

\subsection{Re-calibrated Contrastive Loss}
The contrastive loss has been used fundamentally in training VLMs. The similarity for the contrastive loss with image features $\bm{I}$ and text features $\bm{T}$ is computed as:
\begin{equation}
\label{eq:logits}
    \bm{S} = \bm{I} \cdot \bm{T}^\top,
\end{equation}
where $\bm{I}$,$\bm{T}\in\mathbb{R}^{B \times D}$ and $\bm{S} \in \mathbb{R}^{B \times B}$. $B$ and $\mathit{D}$ represent the batch size and dimension of the feature vector.
To compute the loss from the similarity, we sum the cross-entropy terms for image-text $\mathcal{L}_{ce}^{I, T}$ and text-image $\mathcal{L}_{ce}^{T, I}$ contrastive matches as: 
\begin{equation}
\label{eq:clip-loss}
\begin{gathered}
\mathcal{L} = (\mathcal{L}_{ce}^{I,T} + \mathcal{L}_{ce}^{T,I}) / 2, \\
\mathcal{L}_{ce}^{I,T} = - \sum_{i=1}^{B} \log \sigma^{I}(i, i), \quad \mathcal{L}_{ce}^{T,I} = - \sum_{j=1}^{B} \log \sigma^{T}(j, j), \\
\end{gathered}
\end{equation}
where the terms $\sigma^{I}\!(i,i)$ and $\sigma^{T}\!(j,j)$ denote the probabilities of the diagonal matches. Specifically, $\sigma^{I}\!(i,i)$ represents the probability that the $i$-th image is paired with its symmetric $i$-th text prompt, while $\sigma^{T}\!(j,j)$ is the probability that the $j$-th text prompt is paired with the symmetric $j$-th image.
To compute the probabilities, the raw similarity scores from matrix $\bm{S}$ must be normalized by applying a softmax function to the similarities as follows:
\begin{equation}
\label{eq:similarity}
\begin{gathered}
\sigma^{I}\!(x,\!y)\!=\!\frac{\exp(\bm{s}_{x,y})}{\sum_{j=1}^{B} \exp(\bm{s}_{x,j})}, \quad
\sigma^{T}\!(x,\!y)\!=\!\frac{\exp(\bm{s}_{x,y})}{\sum_{i=1}^{B} \exp(\bm{s}_{i, y})},
\end{gathered}    
\end{equation}
where $\bm{s}$ denotes a single similarity value and the first and second subscripts of $\bm{s}$ represent the image and text prompt indices of the similarity matrix $\bm{S}$. The text-image contrastive loss $\mathcal{L}_{ce}^{T, I}$ is computed by permuting the indices to normalize similarity elements across text prompt data. 
This cross-entropy of Equation~\ref{eq:clip-loss} assumes the unsupervised setting where the class labels are not utilized. Therefore, an image sample with its respective text prompt will only be used as a single positive pair, and all others will be negative pairs, as shown in \cref{fig:loss-clip}.

In contrast, an image sample with a different sample's text prompt might be considered as a positive pair, as shown in \cref{fig:loss-softce} when there are multiple same-class samples in a batch.
We therefore extend the $\mathcal{L}_{ce}^{I,T}$ to the supervised setting as:
\begin{equation}
\label{eq:softce-i2t}
\mathcal{L}_{softce}^{I,T} = - \sum_{i=1}^B \frac{1}{|\bm{P}(i)|} \sum_{k \in \bm{P}(i)} \log \sigma^{I}(i, k),
\end{equation}
where $\mathcal{L}_{softce}^{I,T}$ is the soft cross-entropy loss and
$\bm{P}(i) = \{k|y_k=y_i\}$ denotes the set of indices whose labels match $y_i$. This soft-target cross-entropy is widely used in conjunction with MixUp \citep{zhang2018mixup} or CutMix \citep{yun2019cutmix}. 
This objective is also the cross-modal form of SupCon~\citep{khosla2020supervised} because image and text anchors belong to different modalities and therefore have no self-term to exclude. Under a uniform target, each positive receives a target weight of $\nicefrac{1}{|\bm{P}(i)|}$, so its gradient contribution decreases as the number of same-class positives increases.
To extend this multi-positive objective to the augmented contrastive space, our loss is formulated by considering the concatenated similarity matrices of the clean and augmented data as:
\begin{equation}
\label{eq:our-loss}
\mathcal{L}_{ours} = (\mathcal{L}_{ours}^{[I I'],[T T']} + \mathcal{L}_{ours}^{[T T'],[I I']} ) / 2,
\end{equation}
where $[I I'],[T T']$ denote the concatenated clean and augmented image and text features, respectively.
The expanded contrastive space increases the number of valid positive image-text matches per anchor, exacerbating positive-gradient dilution under the uniform target. We therefore introduce loss re-calibration to mitigate this effect.
We adjust the loss value as:
\begin{equation}
\label{eq:ours}
    \mathcal{L}_{ours}^{[I I'],[T T']} = -\sum_{i=1}^B \frac{1}{|\bm{P}(i)|} \sum_{k \in \bm{P}(i)} \log \left\{\sigma^{I}(i, k) \cdot \psi(i)\right\},
\end{equation}
where $\psi$ denotes the adaptive re-calibration factor.
Expanding the logarithm separates it from the normalized softmax term as:
\begin{equation}
\label{eq:ours-clean}
\mathcal{L}_{ours}^{[II'],[TT']} = \mathcal{L}_{softce}^{I,T} - \sum_{i=1}^{B}\log\psi(i),
\end{equation}
where $\sigma^{I} (i, k)$ remains a valid normalized probability, and $-\log\psi(i)$ acts as an adaptive margin.
We set $\psi$ to be adaptive according to the number of positive and negative pairs of a given sample $i$ as:
\begin{equation}
\label{eq:re-calibration-factor}
    \psi(i) = \displaystyle\frac{\sum_{k\in\bm{P}(i)} \exp(\bm{s}_{i,k})}{\sum_{t\in\bm{N}(i)} \exp(\bm{s}_{i,t})},
\end{equation}
where $\bm{s}_{i,j}$ is the logit between image anchor $i$ and text $j$, and $\bm{N}(i) = \{t|y_t \neq y_i\}$ is the set of negative indices.
Equation~\ref{eq:re-calibration-factor} measures the relative similarity
between the positive and negative distributions for each sample. The numerator
aggregates the exponential similarities over all positive pairs and grows with
$|\bm{P}(i)|$, while the denominator aggregates those over all negative pairs.
The ratio thus quantifies how strongly positive evidence dominates negative
interference for anchor $i$.
Excluding positive pairs from the denominator is intentional. With a union
denominator comprising both positive and negative pairs, $\psi(i)$ reduces to
the probability mass that $\sigma^{I}$ assigns to the positive set. It is then
bounded within $(0,1)$ and approaches $1$ as positive evidence dominates, so
the adaptive margin $-\log\psi(i)$ vanishes and merely duplicates the signal of
$\mathcal{L}_{softce}^{I,T}$. A denominator consisting solely of negative pairs
leaves $\psi(i)$ unbounded above, and the margin instead grows with the gap
between positive evidence and negative interference. Furthermore, because
$\psi(i)$ is independent of $k$, the margin in Equation~\ref{eq:ours-clean} is
not divided by $|\bm{P}(i)|$. Unlike the soft-target term, it retains full
strength as the number of positives increases.
Since $\psi(i)$ is unbounded above, $-\log\psi(i)$ can be negative. The
objective nevertheless remains bounded below under CLIP's standard constraints.
CLIP uses $L_2$-normalized features and caps the logit scale at $100$, so
$\bm{s}_{i,j}\in[-\tau,\tau]$ for finite $\tau$ and
\begin{equation}
\log\psi(i) \in \left[
\log\frac{|\bm{P}(i)|}{|\bm{N}(i)|}-2\tau,\;
\log\frac{|\bm{P}(i)|}{|\bm{N}(i)|}+2\tau
\right].
\end{equation}
As $|\bm{P}(i)|/|\bm{N}(i)|$ is bounded by the size of the concatenated batch,
the adaptive margin in Equation~\ref{eq:ours-clean} has a finite lower bound. A
value of $\psi(i)>1$ indicates dominant positive evidence rather than an
invalid probability.
The objective therefore strengthens the learning signal as the number of
positives increases, mitigating positive-gradient dilution. Algorithm~1
provides PyTorch-like pseudocode. Related studies have explored adaptive loss weighting and margin-based objectives~\citep{deng2019arcface, lin2017focalloss}.

\begin{table}[t]
\label{alg:example}
    \centering
    \begin{tabular}{p{0.94\linewidth} r}
    \toprule
    \textit{Algorithm 1.} PyTorch-like pseudocode of the proposed re-calibrated contrastive loss function.\\
    \midrule
    \begin{minipage}[t]{0.94\linewidth}
    \begin{lstlisting}[style=alg]
"""
similarity_i: Similarities of an image to text # 2B, 2B
similarity_t: Similarities of text to image # 2B, 2B
targets: ground-truth of positive pairs # 2B, 2B
B: batch size
The comment in the middle of the code specifies the shape 
when computing the loss of similarity_i and targets.
"""
# compute: adaptive re-calibration factor
def recal(similarities, targets, eps=1e-8):
    exp_sim = torch.exp(similarities)
    pos = torch.sum(exp_sim * targets, dim=-1)
    neg = torch.sum(exp_sim * (1 - targets), dim=-1)
    return pos / (neg + eps)

# compute: our re-calibrated contrastive loss
def loss(similarities, targets, eps=1e-8):
    prob = torch.softmax(similarities, dim=-1)
    prob = torch.log(prob * recal(similarities, targets, eps) + eps)
    each_loss = torch.sum(-targets * prob, dim=-1)
                / (torch.sum(targets, dim=-1) + eps)
    return torch.mean(each_loss)

# compute: total re-calibrated contrastive loss
def loss_t(similarity_i, similarity_t, targets, eps=1e-8):
    return (loss(similarity_i, targets, eps)
            + loss(similarity_t, targets, eps)) / 2
    \end{lstlisting}
    \end{minipage} \\
    \bottomrule
    \end{tabular}
\end{table}

\subsection{Multi-sample Fusion}
Augmentation methods, such as MixUp~\citep{zhang2018mixup} and CutMix~\citep{yun2019cutmix}, combine multiple images into a new training sample.
However, applying a similar augmentation approach to text prompts presents challenges due to the ambiguity of merging multiple distinct class words into a cohesive format.
To address this, rather than merging text data at the prompt level, we adopt a feature-level fusion approach, similar to \citet{verma2019manifoldmixup}, which can effectively be applied to both image and text data.
We propose multi-sample fusion on both image and text features.

Let $W^I$ and $W^T$ denote the projection matrices for the image and text encoders, respectively. We define $X^I = \{x^I_1, x^I_2, \dots, x^I_B\}$ and $X^T = \{x^T_1, x^T_2, \dots, x^T_B\}$ as the output features from the respective encoders for a mini-batch of size $B$.
The final projected image features $I$ and text features $T$ are represented as $I = W^IX^I$ and $T = W^TX^T$.
To adapt the multi-sample fusion before the projection, the fused feature for the $i$-th sample in the batch is defined as:
\begin{equation}
\label{eq:multi-sample}
\hat{x}^m_i = \lambda x^m_i + (1-\lambda) x^m_{B-i+1},
\end{equation}
where $m \in \{ I, T \}$ represents the modality, $\lambda \in [0,1]$ is a randomly chosen fusion ratio, and $\hat{x}^m_i$ denotes the fused feature. In Equation~\ref{eq:multi-sample}, we pair the $i$-th sample with the ($B - i + 1$)-th sample in the batch. The augmented image and text features in the projected contrastive space are subsequently given by:
\begin{equation}
\label{eq:augmented-features}
\hat{I} = W^I \hat{X}^I, \quad \hat{T} = W^T \hat{X}^T,
\end{equation}
where $\hat{X}^I = \{\hat{x}^I_1, \hat{x}^I_2, \dots, \hat{x}^I_B\}$ and $\hat{X}^T = \{\hat{x}^T_1, \hat{x}^T_2, \dots, \hat{x}^T_B\}$. We use these
projected features to compute the re-calibrated contrastive loss.

\section{Experiments}
We evaluate the proposed method on diverse VLM transfer learning tasks.
Specifically, we examine its effectiveness under distribution shift, extended training, standard transfer learning, and few-shot learning.
We further provide ablation studies and analyses to validate the importance of transformation-aware prompt conditioning, the re-calibrated contrastive loss, and the augmented contrastive space.

\paragraph{Experimental setup.}
We conduct experiments using pre-trained CLIP~\citep{radford2021clip} models.
We use ViT-B/16~\citep{dosovitskiy2020vit} and ViT-B/32 as backbones for distribution shift and transfer learning experiments, and ResNet50~\citep{he2016resnet} for few-shot learning experiments.
For fair comparison, we follow most hyperparameter settings from a previous transfer learning method~\citep{wortsman2022wiseft}.
Detailed hyperparameter settings are provided in the Supplementary Material.
Unless otherwise specified, we fully fine-tune the CLIP model by updating the entire network.
For parameter-efficient transfer learning, we freeze the backbone and update only a single additional multi-layer perceptron (MLP) placed at the end of each image and text encoder.

\subsection{Distribution Shift}
\label{sec:distribution-shift}
We evaluate the proposed method under distribution shift using in-distribution (ID) datasets, including ImageNet~\citep{deng2009imagenet} and iWildCam~\citep{beery2020iwildcam}, together with their corresponding out-of-distribution (OOD) datasets.
We fine-tune CLIP models only on the ID datasets and report performance on both ID and OOD datasets.
For OOD evaluation, we use ImageNet-R~\citep{hendrycks2021imagenetr}, ImageNet-A~\citep{hendrycks2021imageneta}, ImageNet-V2~\citep{recht2019imagenetv2}, ImageNet-Sketch~\citep{wang2019imagenetsketch}, and ObjectNet~\citep{barbu2019objectnet}.
Following previous studies~\citep{wortsman2022wiseft, zhu2023ape}, we report accuracy (\%) except iWildCam, for which we report macro F1~(\%).

\tabDistributionShift

\paragraph{Comparison with state-of-the-art methods.}
We compare our method with prior state-of-the-art approaches, including LinearProb~\citep{radford2021clip}, FFT~\citep{radford2021clip}, LP-FT~\citep{kumar2022lpft}, WiseFT~\citep{wortsman2022wiseft}, FLYP~\citep{goyal2023flyp}, ARF~\citep{han2024arf}, Lipsum-FT~\citep{nam2024lipsumft}, and RAda-FT~\citep{chen2025radaft}.
As shown in \cref{tab:distribution-shift}, our method improves ID performance and achieves the best OOD performance.
In particular, it improves OOD performance by 0.9\% over the previous state-of-the-art method without sacrificing ID accuracy, which remains the highest at 83.3\%.
Across all ImageNet-based datasets, our method improves the harmonic mean of ID and OOD performance by 0.7\% over the previous state of the art.
On iWildCam, it improves the harmonic mean of macro F1 score by 1.4\%.

Notably, our parameter-efficient variant, denoted as Ours (PETL), achieves competitive performance compared with full backbone update methods such as FFT, LP-FT, and WiseFT on ImageNet, despite updating only a single MLP layer.
This result suggests that our method can learn rich transferable representations with a minimal number of trainable parameters.

\figSaturationEpochwise

\paragraph{Overfitting under extended training.}
\cref{fig:saturation-epochwise} presents overfitting under extended training.
In \cref{fig:saturation-flyp}, the training loss continues to decrease, whereas the evaluation loss first decreases and then increases, indicating typical overfitting.
This trend shows that the model increasingly fits the ID training distribution but gradually loses its ability to generalize to unseen or shifted distributions.
In contrast, our method consistently reduces both training and evaluation losses, as shown in \cref{fig:saturation-ours}.
This result indicates that the transformation-aware prompt conditioning and the re-calibrated contrastive loss regularize training beyond early optimization.

\cref{fig:epochwise-id,fig:epochwise-hmean} illustrate the effect of overfitting on model performance.
When we extend the training duration of several state-of-the-art methods, both ID performance and the harmonic mean decline as training progresses.
This degradation shows that additional optimization can harm transferability when the training objective does not sufficiently regularize the model against distribution-specific patterns.
By contrast, our method consistently improves ID performance while increasing the harmonic mean.
These results indicate that our method remains effective with extended training and avoids the OOD generalization degradation observed in existing approaches.
The proposed method enables longer, more stable fine-tuning, allowing the model to benefit from additional training without sacrificing robustness to distribution shift.
Additional details are provided in the Supplementary Material.

\tabTransferLearning
\figFewShot

\subsection{Transfer Learning}
We evaluate transfer learning performance on diverse downstream datasets.
We compare our method with previous approaches on six datasets: Caltech101~\citep{li2022caltech}, PCam~\citep{veeling2018pcam}, ImageNet~\citep{deng2009imagenet}, Flowers102~\citep{nilsback2008flowers102}, StanfordCars~\citep{krause2013stanfordcars}, and iWildCam~\citep{beery2020iwildcam}.
As shown in \cref{tab:transfer-learning}, our method achieves the best performance on most datasets and improves the average performance by 0.8\% over the previous state of the art.
These results demonstrate that the re-calibrated contrastive loss effectively transfers generalized representations to diverse target domains.

\subsection{Few-shot Learning}
We conduct few-shot learning experiments, as shown in \cref{fig:fewshot}, to evaluate whether the proposed method can adapt a pre-trained model using limited domain-specific data.
This setting represents a practical and important evaluation protocol for real-world transfer learning.
We use 11 datasets: Caltech~\citep{li2022caltech}, Describable Textures Dataset (DTD)~\citep{cimpoi2014dtd}, 102 Category Flower Dataset~\citep{nilsback2008flowers102}, ImageNet~\citep{deng2009imagenet}, UCF101~\citep{soomro2012ucf101}, FGVC-Aircraft Benchmark~\citep{maji2013fgvc}, EuroSAT~\citep{helber2018eurosat}, Food101~\citep{bossard2014food101}, SUN397~\citep{xiao2010sun397}, StanfordCars~\citep{krause2013stanfordcars}, and Oxford-IIIT Pet~\citep{parkhi2012oxfordpets}.
We evaluate five few-shot settings with 1, 2, 4, 8, and 16 shots per dataset while keeping the backbone frozen.
We compare our method with state-of-the-art few-shot adaptation methods, including CLIP-Adapter~\citep{gao2023clipadapter}, CoOp~\citep{zhou2022coop}, Tip~\citep{zhang2021tip}, and APE~\citep{zhu2023ape}.
Our method improves performance by 0.8\% in both the 1-shot and 16-shot settings compared with the previous state of the art.
It also consistently performs well across diverse fine-grained classification datasets, demonstrating stronger generalization than prompt tuning and adapter-based approaches.

\subsection{Ablation Study}
We investigate the importance of transformation-aware prompt conditioning. Because VLMs depend on precise correspondence between visual and textual representations, image augmentations can misalign visual features with their paired text prompts.
We design an ablation study to show that enforcing consistency between the image transformation and its fixed text descriptor during augmentation prevents this misalignment.
We also analyze strategies for constructing the augmented contrastive space and evaluate the re-calibrated contrastive loss, then isolate the objective and components such as the loss and multi-sample fusion.

\tabAugmentationAlignment
\subsubsection{Transformation-Aware Prompt Conditioning}
We evaluate the importance of transformation-level consistency between augmented images and text prompts.
In this experiment, we transfer a ViT-B/32 backbone to ImageNet.
We apply image augmentation using either random crop or RandAug~\citep{cubuk2020randaug}.
Text prompts are evaluated under three conditions: fixed, misaligned, and aligned.
The fixed condition uses text prompts without a transformation descriptor.
The misaligned condition randomly inserts a descriptor for a different transformation into the text prompt.
The aligned condition uses text prompts that correctly describe the augmentation applied to the image.

As shown in \cref{tab:augmentation-alignment},  a mismatched descriptor obtains a 57.3\% harmonic mean under RandAug, below the descriptor-free baseline of 61.6\%, while the correct descriptor reaches 62.4\%. These results show that a
correct transformation descriptor is more effective than descriptor-free or mismatched conditioning for VLM transfer learning.
Comprehensive results are provided in the Supplementary Material.

\tabAblationStudy
\subsubsection{Re-calibrated Contrastive Loss and Space}
We incrementally apply the proposed components, starting from the baseline in \cref{tab:ablation-study}.
We use the ViT-B/32 backbone and evaluate ImageNet ID performance for this ablation study.
The baseline, which uses only image-text matches without adaptive re-calibration, achieves 77.2\% accuracy on the ID dataset and 60.4\% harmonic mean.
Expanding the contrastive space, such as using \itmark+\iptpmark\qqquad or \allmark, improves performance by approximately 1\% compared with learning in a single space, such as \itmark\qqquad or \iptpmark.
This result indicates that richer supervision from augmented pairs improves generalization.

A similar improvement is observed when applying the proposed re-calibrated contrastive loss.
Furthermore, the re-calibrated contrastive loss consistently improves performance by approximately 1\% across all contrastive space configurations.
Together, these results confirm that the augmented contrastive space and re-calibrated contrastive loss provide complementary benefits, improving VLM transfer learning performance by approximately 2\%.

\subsubsection{Objective and Component Isolation}
\tabObjectiveComparison
\cref{tab:objective-comparison} compares the objective baselines and isolates the proposed components. Every configuration already includes transformation-aware prompt conditioning and the augmented contrastive space. In our cross-modal setting, SoftCE is equivalent to canonical SupCon because there is no within-modality self term to exclude. When added to the SoftCE baseline with transformation-aware prompt conditioning and the augmented contrastive space, re-calibration and multi-sample fusion further improve ID accuracy by 1.0\% and H.M. by 0.6\%. The differences in this table therefore reflect only the additional contributions of re-calibration and multi-sample fusion, rather than the total gain of our training framework.

\section{Discussion}
\label{sec:discussion}
This section discusses the strengths and limitations to guide future research directions.
Our method introduces transformation-aware prompt conditioning with image augmentation for VLM transfer learning without the need for large language models.
Because augmentation is an effective regularization strategy, it should also be carefully considered in VLM training. Our approach leverages such augmentations to improve transfer learning performance and reduce overfitting, particularly under extended training settings.
Notably, previous methods show degraded performance on OOD datasets as training duration increases, whereas our method remains more stable.

Our re-calibrated contrastive loss strengthens the learning signal from multiple positive pairs in the augmented space, thereby mitigating positive-gradient dilution.
However, constructing the augmented contrastive space requires clean and transformed data from each training image, increasing training cost. Relative to FLYP, per-step runtime and peak memory are $1.0\times$ at a batch size of 8 and $1.6\times$ at a batch size of 64, respectively. Multi-sample fusion is applied to features only during training, so inference cost is identical to that of the backbone. Detailed analysis is provided in the Supplementary Material.
We therefore argue that enabling sufficiently long training without severe overfitting provides a favorable trade-off.

Moreover, the compatibility of our method with parameter-efficient transfer learning and few-shot scenarios highlights its practicality for real-world applications, where adaptation with limited labeled data is often required.
Looking ahead, it would be valuable to construct a predefined set of augmentation combinations from the large space of possible image and text transformations, following strategies such as AutoAug~\citep{cubuk2019autoaugment} or AugMix~\citep{hendrycks2020augmix}.
These directions may further improve the effectiveness and robustness of Vision-Language Models across diverse transfer learning scenarios.

\section{Conclusion}
This paper presents a practical framework for transfer learning of Vision-Language Models. Our transformation-aware prompt conditioning improves model regularization while preserving transformation-level consistency between image and text modalities. The augmented contrastive space further supports learning from clean, transformed, and cross-paired image-text matches. Our re-calibrated contrastive loss mitigates positive-gradient dilution under a uniform soft target as the number of positives increases, allowing the model to benefit from broader positive exposure. Multi-sample fusion provides additional regularization without introducing prompt-level ambiguity.

Experiments on ImageNet and iWildCam demonstrate strong transferability in both in-distribution and out-of-distribution settings. Our method mitigates overfitting and consistently improves transfer-learning performance across diverse downstream datasets. It also performs effectively on 11 few-shot benchmarks and remains stable under extended training. These findings encourage further investigation of transformation-aware prompt conditioning and loss re-calibration for robust VLM transfer learning.

\paragraph{Acknowledgments.} This research was supported by the National Research Foundation of Korea (NRF), Electronics and Telecommunications Research Institute (ETRI), and Institute of Information \& Communications
Technology Planning \& Evaluation (IITP), funded by the Korean government [26CS1100, Development of Proprietary Physical AI-based Small-scale Computers and Integrated Soft Suits], and the Korean government (MSIT) (RS-2026-25518808, RS-2026-25617480, and IITP-2026-RS-2023-00255968).

\bibliography{ref}

\appendix
\setcounter{table}{5}
\setcounter{figure}{5}

\section{Experiment Details}
\label{sec:sup-experiment-details}
We clarify the hyperparameter settings used in our experiments.
\cref{tab:sup-hyperparameter-default,tab:sup-hyperparameter-transfer,tab:sup-hyperparameter-imagenet,tab:sup-hyperparameter-iwildcam} show the details of hyperparameters for experiments of our transfer learning, distribution shift, and PETL. $^\dag$ denotes PETL, which updates a few parameters, so we employ a large batch size. Few-shot classification closely follows APE, with only the learning rate and weight decay being set differently for each dataset.
The transfer learning task requires approximately 6 hours on four RTX 3090 GPUs, whereas the other tasks use a single GPU. The PyTorch version is $2.2.1+\text{cu}121$, and the Python version is $3.10$.

\SupTabHyperparameter
\SupTabHyperparameterImageNet
\SupTabHyperparameterIWildCam

\section{Additional Diagnostics and Ablations}

\paragraph{Gradient and optimization stability.}
The re-calibrated objective retains a stronger gradient than SoftCE as the number of positives increases, as shown in \cref{tab:sup-gradient-diagnostics}. Across 800 random batches, both the loss and gradients remain finite, including an adversarial setting with substantially more positives than negatives, for which the gradient norm is 0.083. Over three seeds, ImageNet ID accuracy and H.M. are $79.4\%\pm0.1$ and $62.5\%\pm0.1$, respectively.
\SupTabGradientDiagnostics

\paragraph{Training efficiency.}
The additional clean and transformed branches increase training computation, as quantified in \cref{tab:sup-efficiency}. Across the evaluated batch sizes, runtime and peak memory are at most $1.7\times$ and $1.6\times$ those of FLYP, respectively. Inference uses the original encoders and therefore remains unchanged.
\SupTabEfficiency

\vspace{-.25em}
\paragraph{Transformation count.}
We use $k=2$ to follow RandAug standard two-operation policy. As shown in \cref{tab:sup-transformation-count}, performance is stable for $k\in\{2,3\}$.
\SupTabTransformationCount

\vspace{-.25em}
\section{Additional Results}
\label{sec:sup-additional-results}
We present the additional experimental results. \cref{fig:sup-epochwise} illustrates overfitting under extended training on ImageNet and its distribution-shifted variants.
\SupFigEpochwise

\end{document}